\documentclass[twocolumn, switch]{article} 

\usepackage{preprint}

\usepackage{amsmath, amsthm, amssymb, amsfonts}

\usepackage[utf8]{inputenc}	
\usepackage[T1]{fontenc}	
\usepackage{xcolor}		
\usepackage[colorlinks = true,
            linkcolor = purple,
            urlcolor  = blue,
            citecolor = cyan,
            anchorcolor = black]{hyperref}	
\usepackage{booktabs} 		
\usepackage{nicefrac}		
\usepackage{microtype}		
\usepackage{lineno}		
\usepackage{float}			

\usepackage{lipsum}		

\usepackage{newfloat}
\DeclareFloatingEnvironment[name={Supplementary Figure}]{suppfigure}
\usepackage{sidecap}
\sidecaptionvpos{figure}{c}

\usepackage{titlesec}
\titlespacing\section{0pt}{12pt plus 3pt minus 3pt}{1pt plus 1pt minus 1pt}
\titlespacing\subsection{0pt}{10pt plus 3pt minus 3pt}{1pt plus 1pt minus 1pt}
\titlespacing\subsubsection{0pt}{8pt plus 3pt minus 3pt}{1pt plus 1pt minus 1pt}

\usepackage[backend=biber,sorting=none]{biblatex}
\usepackage{algorithm}
\usepackage{algorithmic}
\usepackage[title]{appendix}

\DeclareMathOperator{\sign}{sign}

\title{\textit{How Much} Can I Trust You? --- 
Quantifying Uncertainties in Explaining Neural Networks}

\usepackage{xwatermark}
\newwatermark[firstpage,color=black!90,angle=0,scale=0.28, xpos=-0in,ypos=-5in]{*: Equal contribution}

\author{
    Kirill Bykov$^{*}$\\
    ML Group, TU Berlin, Germany\\
	\texttt{kirill.bykov@campus.tu-berlin.de}\\[1ex]
	\And
	Marina M.-C. H\"ohne$^{*}$\\
    ML Group, TU Berlin, Germany\\
	\texttt{marina.hoehne@tu-berlin.de}\\[1ex]
	\And
	Klaus-Robert M\"uller\\
	ML Group, TU Berlin, Germany \\
	MPII and Korea University, Seoul\\
	Google Research, Brain Team, Berlin\\
    \texttt{klaus-robert.mueller@tu-berlin.de} \\[1ex]
    \And
	Shinichi Nakajima\\
	ML Group, TU Berlin, Germany \\
	RIKEN AIP, Tokyo, Japan \\
	\texttt{nakajima@tu-berlin.de} \\[1ex]
	\And
	Marius Kloft\\
	Department of Computer Science, 
	TU Kaiserslautern, Germany\\
	\texttt{kloft@cs.uni-kl.de} 
	}
\date{}

\begin{document}
\twocolumn[
  \begin{@twocolumnfalse}
\maketitle
\begin{abstract}
Explainable AI (XAI) aims to provide interpretations for predictions made by learning machines, such as deep neural networks, 
in order to make the machines more transparent for the user and furthermore trustworthy also for applications in e.g. safety-critical areas.
So far, however, no methods for \textit{quantifying uncertainties of explanations} have been conceived, which is  problematic in domains where a high confidence in explanations is a prerequisite. 
We therefore contribute by proposing a new framework that allows to convert \textit{any} arbitrary explanation method for neural networks into an explanation method for Bayesian neural networks, with an in-built modeling of uncertainties. 
Within the Bayesian framework a network's weights follow a distribution that extends standard single explanation scores and heatmaps to distributions thereof, in this manner translating the intrinsic network model uncertainties into a quantification of explanation uncertainties.  
This allows us for the first time to carve out uncertainties associated with a model explanation and subsequently gauge the appropriate level of explanation confidence for a user (using percentiles).
We demonstrate the effectiveness and usefulness of our approach extensively in various experiments, both qualitatively and quantitatively. 
\end{abstract}
 \vspace{0.5cm}
  \end{@twocolumnfalse}
]
\section{Introduction}
Deep neural networks (DNNs) can learn  highly complex, non-linear predictors that are successfully applied across sciences, humanities and engineering. 
However, in contrast to linear learning machines, DNNs are unable to directly reveal their prediction strategy, which can be a concern in various areas of application, such as safety critical areas or the sciences, where transparency and insight is a must-have. As a countermeasure, the General Data Protection Regulation (GDPR) by the European Union requires from AI systems the transparency of AI-based decision-making systems, to alleviate the risk potential of automated decision-making systems. 

Addressing this challenge, the field of XAI has emerged establishing techniques to explain predictions made by nonlinear learning machines. 
Existing approaches can be categorised into model-agnostic and model-aware explanation methods. Model-agnostic approaches are based on the output of the learner, treating it as a black box \cite{ribeiro2016should,lundberg2017unified,vidovic2016feature}. While being computationally costly, these methods can be widely applied across a large variety of learning methods.
Model-aware methods are specifically crafted for certain types of learners, for instance, feed-forward artificial neural networks \cite{selvaraju2017grad,yosinski2015understanding,bach2015pixel,kindermans2017learning,baehrens2010explain,montavon2018methods,samek2020toward} or various unsupervised learning methods \cite{kauffmann2019clustering,kauffmann2020towards}.

The question of `why should I trust my model' has been discussed intensively by the XAI community \cite{ribeiro2016should,lundberg2017unified,vidovic2016feature,selvaraju2017grad,yosinski2015understanding,bach2015pixel,kindermans2017learning,baehrens2010explain,montavon2018methods,dombrowski2019explanations,samek2020toward}. With the present work, we contribute a more fine-grained analysis, focusing on the question: \textit{how much} can we trust a learning machine (see Figure 1).
To this end, we quantify 
(un)certainties in explanations and visualize them.
Uncertainty generally arises in situations with insufficient information or stochasticity across areas, including physics \cite{heisenberg1949physical}, economics \cite{knight}, and information theory \cite{shannon1948mathematical}.

\begin{figure*}[ht]
\centering
\includegraphics[width=\linewidth]{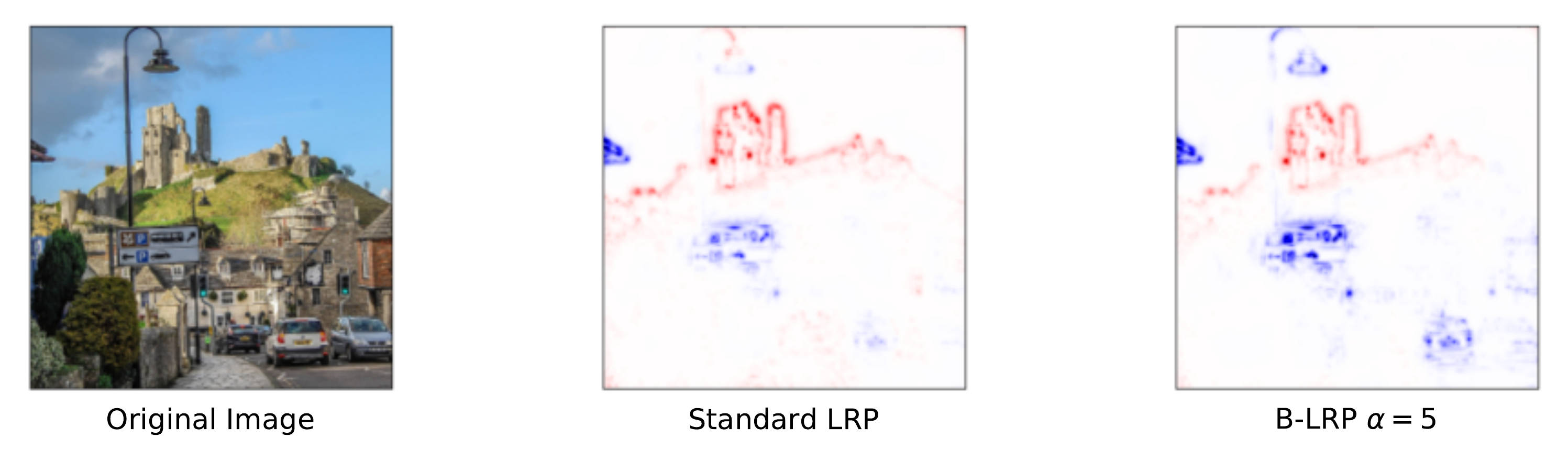}
\vspace{-0.4cm}
\caption{The image shown to the left was classified as \emph{castle} by a VGG16 network pretrained on Imagenet. The heatmap in the center shows the output of LRP, an explanation method for neural networks. LRP correctly determines the castle as relevant (indicated by red color) but gives additional relevance to the gas lamppost. Our proposed B-LRP method can give more cautious explanations, highlighting only the parts of an image where the classifier is the most certain about, here, the castle---not the lamppost (see heatmap to the right, where the gas lamppost has become blue). }
\end{figure*}

Uncertainty is also fundamental to the field of machine learning, e.g. in statistical learning theory (e.g.~\cite{vapnik1995nature,mohri2018foundations}) or Bayesian machine learning (e.g.~\cite{murphy2012machine}). Surprisingly,---as far as we know---there is no
prior work on explaining neural networks that quantifies the inherent uncertainties of explanations. Note that a blind belief in explanations, without taken into account the possibility of uncertainty, can lead to an excessive unwarranted trust in explanations.

\begin{figure*}[!t]
\centering
 \includegraphics[width=0.9 \textwidth]{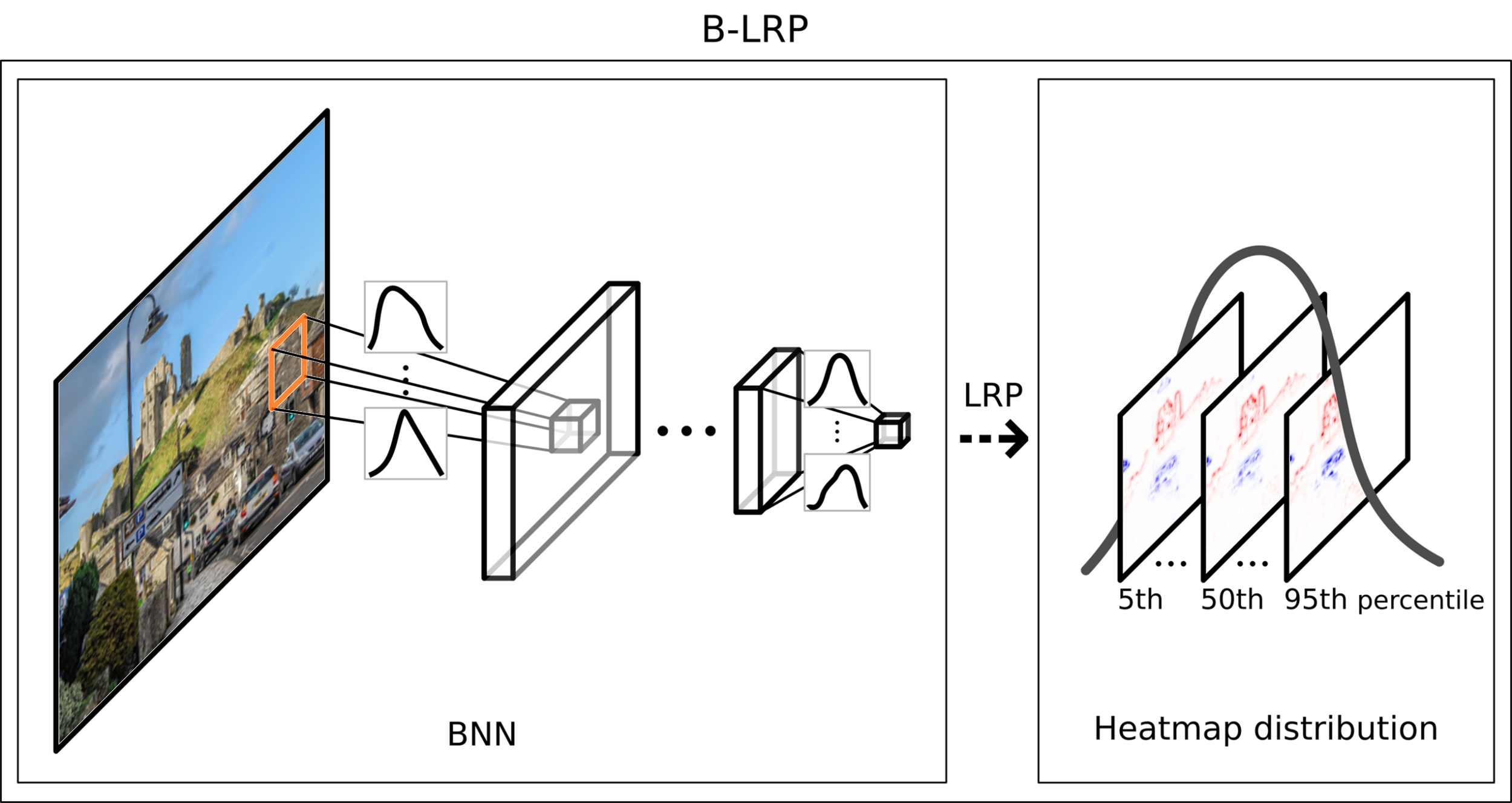}
\caption{Overview of the proposed B-LRP procedure. For a given input, standard LRP generates from a trained neural network an explanation
as a heatmap. Our B-LRP considers a Bayesian neural network (shown to the left), which induces a distribution over neural networks. 
Subsequently, applying LRP evokes a distribution of heatmaps (shown to the right).
The variation in this distribution informs us about the (un)certainty in explanations. B-LRP considers a percentile of the heatmap distribution, leading to more cautious or risky explanations, depending on the chosen percentile.
\label{fig:BLRP}}
\end{figure*}

The core idea of our methodology is the quantification and
visual disclosure of uncertainties in neural network explanations. We propose a knowledge transfer of uncertainties, gained by Bayesian neural networks towards 
the explanation of any neural network---using any  explanation method.
BNNs are well established for assessing uncertainties in neural networks; e.g., interesting contributions to the visualisation of uncertainties in BNNs were made by \cite{kwon2018uncertainty, kendall2017uncertainties}.  
However, there is a lack of prior work on explaining the predictions made by BNNs, which we intend to fill with our present work. 
We would like to reiterate that our novel approach can be applied to any explanation method of neural networks---be it model-agnostic or model-aware---and to any (approximate) inference procedure of BNNs.
The main contributions of this paper are as follows:
\begin{itemize}
\item We propose a new methodology that can leverage any existing explanation method for neural networks to an explanation method for BNNs.
\item We study our approach in detail for a particular explanation method---layer-wise relevance propagation (LRP) \cite{bach2015pixel}---thus proposing the first concrete explanation method of BNNs--called B-LRP.
\item Our B-LRP approach (illustrated in Figure~\ref{fig:BLRP}) provides us with a novel manner of explanation because it outputs a distribution, which we can exploit in interesting ways: 
\begin{enumerate}
\item By considering percentiles of the explanation distribution, we can instantiate more cautious or risky explanations than standard LRP. The choice of the percentile thereby governs the risk. For instance, we observed empirically that the $5$th percentile yields more conservative explanations (i.e., prioritized strategies) than the 50th percentile (which is highly similar to standard LRP). 
\item We can visually describe areas of certainty and uncertainty within any example (e.g., image). 
\item B-LRP reveals the varying importance of the multiple prediction strategies used by the learner. For instance, in Figure 1 (right) the gas lantern is not assigned any positive relevance for the prediction class castle by our proposed method. 
\end{enumerate}
\item Although showcased here for LRP, our proposed methodology for explaining neural networks under uncertainty can in principle be applied to any explanation method for neural networks.
\end{itemize}



We study and demonstrate the validity of the above findings in various experiments 
(from image categorization and handwritten-digit classification as well as digital pathology). 
Qualitative and quantitative experiments nicely underline the usefulness of the B-LRP framework, which we provide as an open source PyTorch implementation\footnote{\url{https://github.com/lapalap/B-LRP}}.

\section{Explaining Bayesian Neural Networks}

In this section, we give background on BNNs and LRP. We then introduce Bayesian LRP, an explanation method for BNNs, which is based on LRP .

\subsection{Bayesian Neural Networks}\label{BNN} 

From a statistical perspective, DNNs are trained using the principle of maximum likelihood estimation (MLE), aiming to minimise a predetermined error function. 
Although the MLE procedure is efficient since networks learn only a fixed set of weights, vanilla networks suffer from the inability of specifying uncertainties on the learned weights and subsequently on the prediction.
In contrast, Bayesian neural networks (BNNs) estimate the posterior \textit{distribution} of weights, and thus, provide uncertainty information on the prediction.
Particularly, in critical real world applications of deep learning---for instance, medicine \cite{hagele2020resolving,holzinger2019causability,klauschen2018scoring} and autonomous driving \cite{koo2015did,wiegand2019drive}---where  predictions are to be highly precise and wrong predictions could easily be fatal, the availability of uncertainties of predictions can be of fundamental advantage.



Let $f(\cdot;W): \mathbb{R}^d \rightarrow \mathbb{R}^k$ be a general feedforward neural network with the weight parameter
$W \subset \mathcal{W}$.
The conditional output distribution, given an input $x \in \mathbb{R}^d$ and the parameter $W$, is typically modeled as
$p(y |f(x; W)) = \mathrm{Multinomial}_k (y; f(x; W))$ for classification,
or $p(y |f(x; W)) = \mathrm{Gauss}_k (y; f(x; W), \Sigma)$ for regression with Gaussian noise. Given a training dataset $\mathcal{D}_{\mathrm{tr}} = \{{x}_n, y_n \}_{n=1}^N$, 
%
Bayesian learning (approximately) computes
the posterior distribution 
\begin{align}
p\left(W |\mathcal{D_{\mathrm{tr}}}\right)
= \textstyle
\frac{p(\mathcal{D_{\mathrm{tr}}} | W) p(W)}
{  \int_{\mathcal{W}} p(\mathcal{D_{\mathrm{tr}}} | W) p(W) dW} ,
\label{eq:BayesPosterior}
\end{align}
where $p(W)$ is the prior distribution of the weight parameter.
After training,
the output for a given test sample is predicted by the predictive distribution:
\begin{align}
p(y | x,\mathcal{D_{\mathrm{tr}}})
&=\textstyle \int_{\mathcal{W}} p(y |f(x; W))
p(W | \mathcal{D_{\mathrm{tr}}}) dW.
\label{eq:BayesPredictive}
\end{align}
Since the denominator of the posterior, shown in Eq.~\eqref{eq:BayesPosterior}, is intractable for neural networks, many approximation methods have been proposed, e.g., Laplace approximation \cite{Ritter18}, variational inference \cite{Graves11,Osawa19}, MC dropout \cite{Gal16}, variational dropout \cite{Kingma15,Molchanov17}, and MCMC sampling \cite{Wenzel20}.  Such approximation methods support efficient sampling from the approximate posterior, which allows us to obtain output samples from the  predictive distribution, given in Eq.~\eqref{eq:BayesPredictive}, for uncertainty evaluation along with prediction for a test sample $x$.
\subsection{Layer-wise Relevance Propagation (LRP)}
%
Many explanation methods
attribute \emph{relevance} to
the input or intermediate nodes  \cite{bach2015pixel,vidovic2016feature,selvaraju2017grad,montavon2018methods}.
A node with a high relevance is supposed to be responsible for the output---%
prediction will be affected much if the node is blocked/removed. 
In this paper, we demonstrate our new explanation framework
based on Layer-wise Relevance Propagation (LRP),
one of the most popular approaches
that backpropagate the relevance from the output 
backwards through  the layers of the network.
However, our proposed framework is general and thus applicable to any explanation method
that attributes relevance scores.
%
In each layer, LRP redistributes the relevance of the output nodes to the input nodes, based on how much each input node affects the output node.
The standard LRP-0 method 
uses a rule that distributes the relevance
in proportion to the contributions of each input to the neuron activation:
$R_i^{(l)} = \sum_j \frac{z_{ij}^{(l+1)}}{\sum_{i'}z_{i'j}^{(l+1)}} R_j^{(l+1)},$
where $z_{ij}^{(l+1)} = x_i^{(l)}w_{ij}^{(l,l+1)}$
is the activation computed in the forward pass at the $(l+1)$-th layer and $\{w_{ij}^{(l,l+1)}\}$ are the learned weight parameters of the network. 

In our work we use the best practice LRP rule, namely LRP-CMP, which was recently published by Kohlbrenner et al. \cite{kohlbrenner2019towards,samek2020toward} and LRP-$\varepsilon$ rule \cite{bach2015pixel}. LRP-CMP uses different basic LRP rules, i.e., LRP-0 rule, LRP-$\varepsilon$ rule, and LRP-$\gamma$ \cite{bach2015pixel} for different layer types of the deep neural network and thus acts like an enhanced combination of those.

\subsection{Relevance Distribution and Bayesian LRP (B-LRP)}
\begin{figure*}[!h]
\centering
 \includegraphics[width=\textwidth]{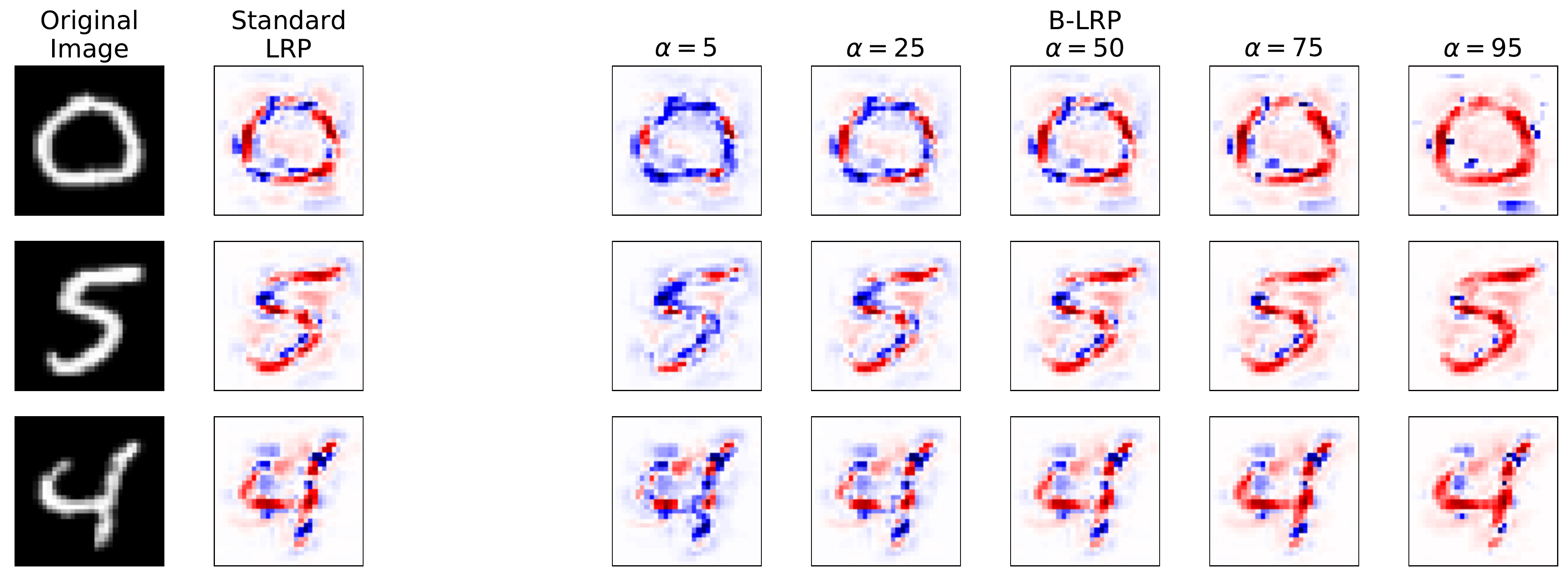}
\vspace{-0.4cm}
\caption{Exemplary explanations of handwritten digits (red/blue indicates positive/negative relevance).
Each row corresponds to a particular input. From left to right: original image, standard LRP explanation,
and proposed B-LRP explanation using the 5, 25, 50, 75, and 95th percentile.
}
\label{fig:expMnist}
\end{figure*}

\label{distribution of explanations} 

Let us denote the relevance of an input $x$ by $R(x; W)\in\mathbb R^d$.
The relevance depends on the learned parameter value $W$, which---for BNNs---follows a posterior distribution.
Therefore, we can naturally consider the distribution of the relevance induced by the BNN.
Let $g(x; W)$ be a vector that contains information necessary to compute the relevance, i.e., there exists a function $\widetilde{R}(x, \cdot)$ such that
$R(x, W) = \widetilde{R}(x, g(x; W))$. 
In the case of LRP-$\gamma$ \cite{montavon2019layer} for ReLU activation, 
$g(x; W)$ can be explicitly written 
as the derivative of the output with respect to the input,
\begin{align}
g(x; W)
&=\textstyle \frac{\partial f(x; W)}{\partial x} 
\notag,
\qquad
\mbox{because}\\
R(x, W) &=\textstyle \widetilde{R}(x, g(x; W))
= x \odot \frac{\partial f(x; W)}{\partial x}.
\notag
\end{align}
Here $\odot$ is the element-wise product.
For general explanation methods, $g(x; W)$ should contain inputs/outputs of all  layers that are used for relevance computation.

Given a posterior distribution on $W$, 
we can define the distribution of relevance as
\begin{align}
p(R | x, \mathcal{D_{\mathrm{tr}}})
&=\textstyle \int  \left\{ \int p(R |x, g) p(g |x, W)dg
\right\}
p(W | \mathcal{D_{\mathrm{tr}}})  dW,
\label{eq:RelevanceDistribution}
\end{align}
where the inner integral in the curly brackets
describes how the relevance is attributed through $g$, given $x$ and $W$. 
When we consider only $W$ as a random variable, $R(x, W)$ is a deterministic mapping from the input and the weight parameter spaces to the relevance space.
In this case, 
the inner integral is simply given by the Dirac measure, i.e., 
$\int
p(R |x, g) p(g |x, W) dg
= \delta \left( R(x; W) \right)
$,
and relevance samples can be obtained 
by computing the relevance for posterior parameter samples: 
\begin{align}
R(x; W) \sim p(R | x, \mathcal{D_{\mathrm{tr}}})
\qquad
\mbox{if}
\qquad
W \sim p(W | \mathcal{D_{\mathrm{tr}}}).
\label{eq:RelevanceSamples}
\end{align}
For efficient sampling,
we typically need to replace the posterior
$p(W | \mathcal{D_{\mathrm{tr}}})$
with its approximation.

Our main proposal, which we call \emph{Bayesian LRP} (B-LRP; illustrated in Figure~\ref{fig:BLRP}), is to treat the relevance of a BNN as a random variable that follows Eq.\eqref{eq:RelevanceDistribution},
and use it to explain the network, with uncertainty information taken into account.
Let $\{R_m\}_{m=1}^M$ be samples from the relevance distribution,
obtained by Eq.\eqref{eq:RelevanceSamples}.
Then we define our B-LRP as follows:
\begin{align}
\textrm{B-LRP}_{\alpha} (x; W)
&=
\mathcal{P}_\alpha\left( \{ R_m\}\right),
\label{eq:B-LRP}
\end{align}
where $\mathcal{P}_\alpha (\{R_m\})$ is an operator computing 
the entry-wise (pixel-wise) percentile from the set $\{R_m\}$ of random samples. B-LRP reveals the pixels where the explanation is uncertain:  for instance, if a pixel has a positive relevance in the 5th percentile, it will be positive in 95\% of the explanations. This means there is strong evidence that it is relevant.
We will demonstrate its usefulness in Section~\ref{sec:Experiments}.


Naturally, Bayesian LRP can be applied only to the networks trained by Bayesian learning.
However, some approximate Bayesian learning 
can be performed as a post-processing applied to  pretrained non-Bayesian networks, which can broaden
the applicability of Bayesian LRP.
A simple and most general way is to apply Laplace approximation \cite{Ritter18}, which is the Gaussian approximation at the MLE solution (assuming a flat prior on $W$).
Given a pretrained NN, we can simply compute the curvature of the log-likelihood to estimate the posterior covariance.
Another, even simpler method is MC dropout \cite{Gal16},
which can be applied to any non-Bayesian network trained with the dropout procedure.
The dropout process can be interpreted as multiplicative noise on the parameter.
Therefore, dropout training can be seen as variational inference with the variational distribution restricted to two-component mixture distributions.
MC dropout can be performed simply by turning on the dropout procedure in the test phase and taking the output random samples as the prediction from networks with the weight parameter sampled from the approximate posterior.







These post-processing procedures for Bayesianizing non-Bayesian learning machines 
do not only broaden the applicability in terms of models, but also offer another use of Bayesian LRP.
Specifically, we can view standard LRP as a statistic that behaves similarly to the median and mean.  From this view, one can assess the reliability/uncertainty of standard LRP by using Bayesian LRP.
In this manner BNNs can make the uncertainty of explanation apparent for any  existing explanation method and even for non-Bayesian learning machines.

\section{Experiments}

\begin{figure*}[!t]
\centering
\begin{minipage}{0.49\linewidth}
\centering
\includegraphics[width=0.8\linewidth]{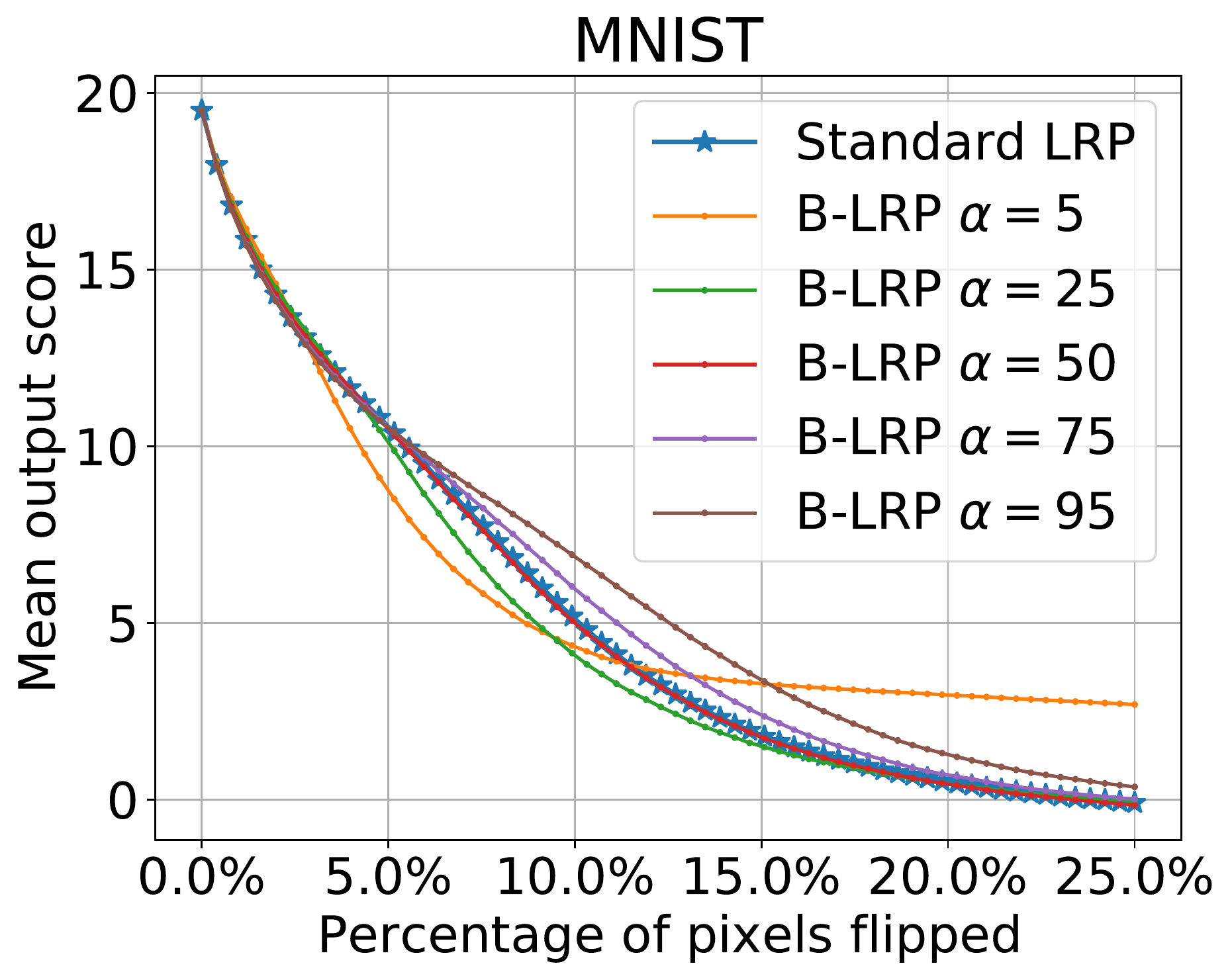}\\
\end{minipage}
\begin{minipage}{0.49\linewidth}
\centering
\includegraphics[width=0.78\linewidth]{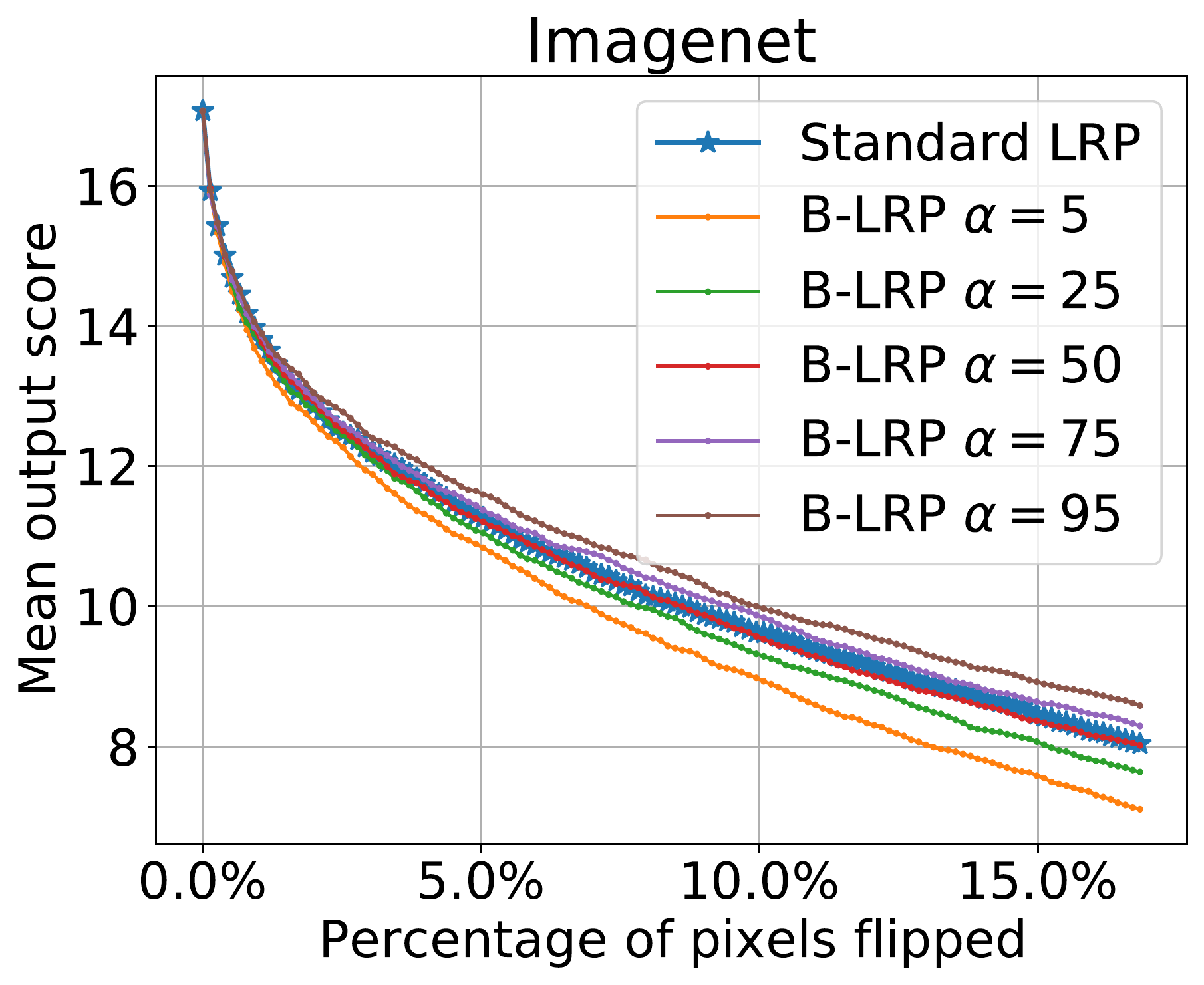}\\
\end{minipage}
\caption{
Quantitative evaluation using the pixel-flipping score on MNIST (left) and ImageNet (right). 
The mean output score for a true class drops when a proportion (horizontal axis) of most relevant pixels are deleted from image and filled with a random value.
The result implies that conservative strategies (5th and 25th percentiles) identify more of the most relevant pixels than standard LRP, especially for the first 12\% of flipped pixels (depends on the domain).
\label{fig:expPixelFlippingMNIST}}
\end{figure*}

\label{sec:Experiments}
In this section we demonstrate the usefulness of our proposed B-LRP method using different LRP rules (LRP-$\varepsilon$ and LRP-CMP rule) on various datasets.
In all experiments we used deep neural networks with dropout layers, trained in a standard non-Bayesian fashion. This demonstrates that our approach is accessible also to users without Bayesian background and can be applied out of the box, without network re-training.
We use \emph{pixel-flipping} \cite{samek2016evaluating} as a quantitative performance criterion, and the standard heatmap normalization \cite{bach2015pixel} for visualization (see Supplementary).

%
\subsection{MNIST}
We first applied the proposed B-LRP method based on the LRP-$\varepsilon$ rule ($\varepsilon = 10^{-9}$) to the MNIST handwritten digits classification dataset. We employ an architecture similar to LeNet \cite{lecun1998gradient} with two additional 
dropout layers (architecture is shown in the supplement),
and applied MC dropout.
As data pre-processing, we resize the handwritten images from $(28 \times 28)$ to  $(32 \times 32)$ pixels and standardize each pixel to mean 0 and standard deviation 1. 
Our proposed B-LRP method allows us to access statistics of the relevance distribution. We consider the
$5$, $25$, $50$, $75$, and $95$th percentiles
and compare them with the
standard (non-Bayesian) LRP-$\varepsilon$ rule in Figure~\ref{fig:expMnist}.
%
We can observe that the median ($50$th percentile) 
is hardly distinguished from standard LRP, while other percentiles give significantly different heatmaps:
the 5-th percentile of B-LRP highlights only pixels of relevance and certainty,
while the 95-th percentile considers many potentially relevant pixels, even those of high uncertainty.
Other examples are shown in the supplement. 


Figure~\ref{fig:expPixelFlippingMNIST} (left) shows  quantitative evaluation results based on pixel-flipping of $1000$ random images from MNIST testset.
We observe that the $50$th percentile behaves similarly to
standard LRP,
while the conservative strategies ($5$th and $25$th percentiles)
identify the most relevant pixels more accurately.
This implies that the relevance map produced by standard LRP is noisy enough to make less relevant pixels happen to get higher relevance scores than the most relevant pixels.
On the other hand, the conservative strategies are less sensitive than standard LRP to find moderately relevant pixels.

\subsection{Imagenet}

\begin{figure*}[!h]

\includegraphics[width= \textwidth]{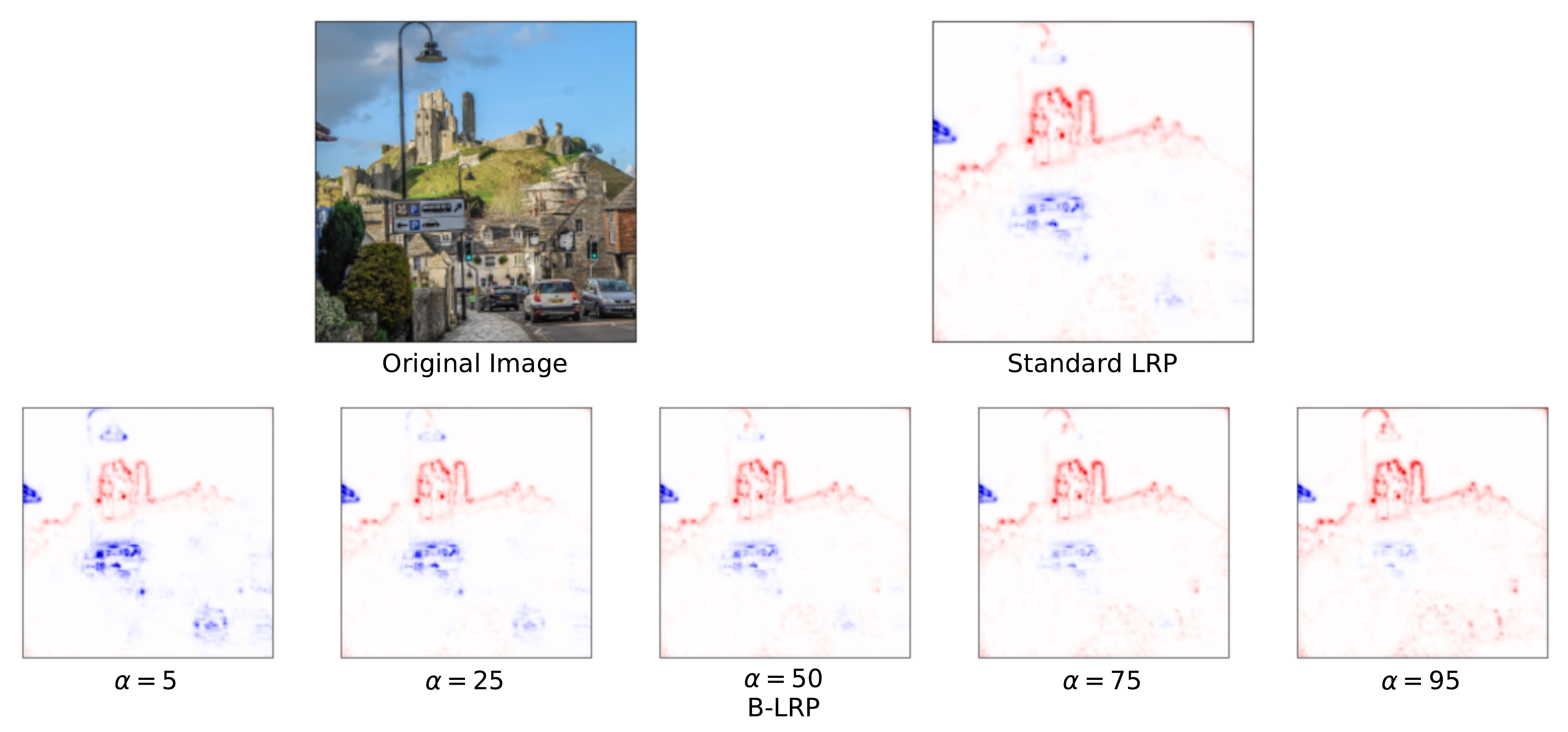}
\vspace{-0.5cm}

\caption{Exemplary explanation of the VGG16 network pre-trained on ImageNet for a "castle" image. 
\textsc{Top row}: Original image and standard LRP explanation. 
\textsc{Bottom}: Bayesian LRP explanation
using the 5, 25, 50, 75 and 95th percentiles. 
We observe that standard LRP attributes positive relevance to pixels outside the area where the castle is seen,
while our Bayesian LRP identifies the relevance of those pixels which are not certain---the color turns red to blue within the credible interval (between 5th and 95th percentiles), meaning that it  crosses the zero relevance.
\label{fig:expImagenet}}
\end{figure*}

Next, we demonstrate the usefulness of the proposed method in analyzing the widely used VGG16 network \cite{simonyan2014very} pre-trained on the Imagenet dataset.
Similarly to the 
previous experiment, we use MC Dropout and
compare our B-LRP using the
5, 25, 50, 75, and 95th percentiles
with standard LRP, visually using LRP-CMP and quantitatively using LRP-$\varepsilon$ ($\varepsilon = 10^{-9}$).
Figure~\ref{fig:expImagenet} shows an example of a ``castle'' image.
We observe that standard LRP-CMP attributes positive relevance to pixels on objects (e.g., lamppost, car) other than the castle.
In contrast, our B-LRP (LRP-CMP based) shows that, at those pixels, the color turns red to blue within the credible interval (between 5th and 95th percentiles), implying that the relevance of those pixels is uncertain.

Figure~\ref{fig:expPixelFlippingMNIST} (right)
shows a quantitative comparison in terms of pixel-flipping scores between standard non-Bayesian LRP-$\varepsilon$ and B-LRP on  300 randomly selected images from Imagenet
from  10 randomly selected classes
(castle, fountain, lemon, llama, pillow, pretzel, teapot, tiger cat, volcano and wine bottle).
For computing percentiles of Bayesian LRP,
we drew $M=100$ parameter samples from the MC dropout posterior
for each image.
Similarly to the MNIST experiment,
we observe that the 5-th percentile of B-LRP identifies the
most relevant pixels more accurately than standard LRP.



\subsection{An illustrative Example: Cancer Pathology}

\begin{figure*}[!t]
\centering

\includegraphics[width= \textwidth]{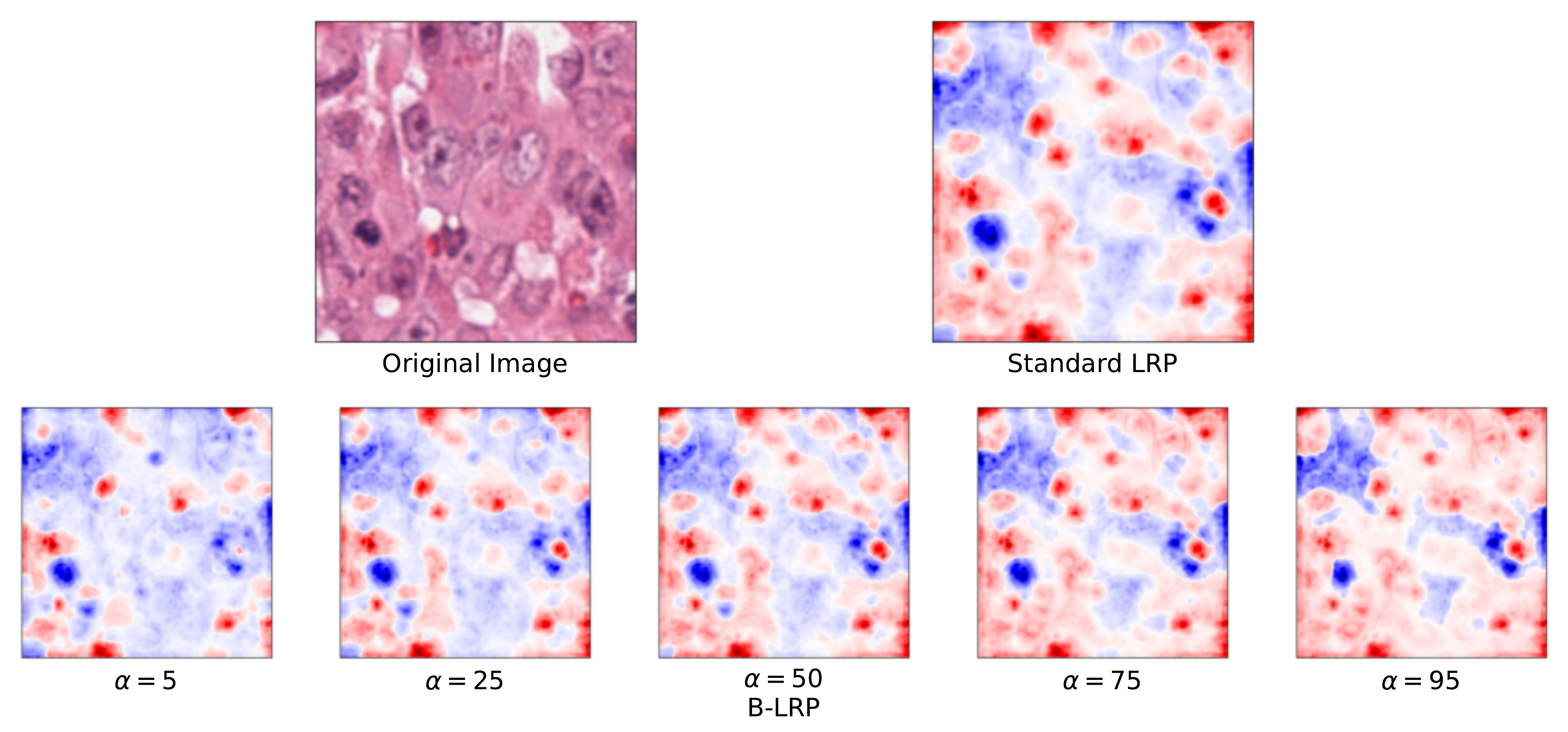}
\vspace{-0.5cm}
\caption{The VGG19 \cite{simonyan2014very} trained on a set of Haematoxylin-eosin-stained Lung adeno carcinoma (LUAD) (and non cancer slices) predicts cancer for the present pathological slice. Standard LRP heatmaps and percentiles using B-LRP are shown indicating different levels of certainty of the explanation heatmaps. Note that the heatmaps have been smoothed by a monotonic tranformation (exact formula could be found in the Appendix) following  \cite{binder2018towards,klauschen2018scoring,hagele2020resolving} to provide an easy to use standard input to a clinician.
\label{fig:cancer_img}}
\end{figure*}

We would like to showcase the usage of our proposed B-LRP method for deep learning in digital pathology. Clearly, the pathology domain requires not only excellent and robust predictions
(here classification cancer vs. non-cancer of haematoxylin-eosin-stained Lung adeno carcinoma (LUAD)\footnote{LUAD dataset can be downloaded from \url{github.com/MiriamHaegele/patch_tcga}}, see \cite{hagele2020resolving}),
but most importantly explanations and insight about why an image was classified as cancerous by the learning machine. 
Here a VGG19 deep learning model provides an out-of-sample prediction for a novel pathological slice (see Fig.\ref{fig:cancer_img}) which it classifies correctly as cancer, along with an LRP heatmap explanation. Our novel B-LRP method allows to additionally provide an estimate of the  explanation uncertainty. 

\begin{figure*}[h]

\includegraphics[width= \textwidth]{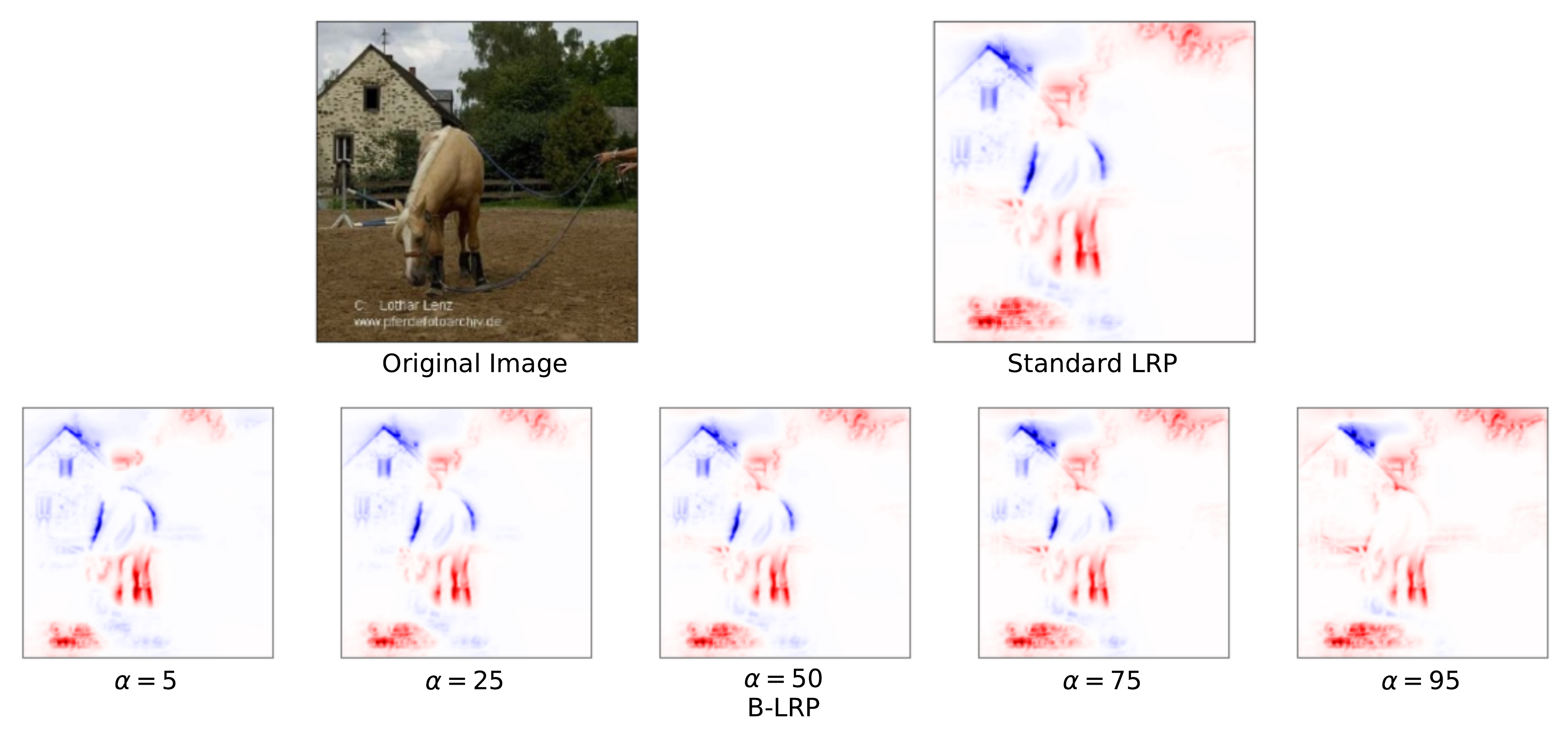}
\vspace{-0.5cm}
\caption{Exemplary visualization of the \emph{clever hans} effect for an image depicting a horse. Bayesian LRP allows us to confirm that the clever Hans effect exists with high certainty, since the watermark in the bottom left is attributed positive relevance consistently within the credible interval.
\label{fig:expCleverHans}}
\end{figure*}

Figure~\ref{fig:cancer_img}  
shows the original tumor image and the explanations provided by standard LRP-CMP and the 5th to 95th percentiles of B-LRP. Inspecting the percentiles provided by B-LRP, we are able 
to show a spectrum ranging from most certain explanations for the trained deep-learning model (5th percentile) to most uncertain ones (95th percentile). Interestingly, the model has captured the underlying areas encompassing cancerous cells (red) and non-cancerous cells (blue) very well and with a high certainty of explanation at the 5th percentile level. Less stringent percentiles allow for more uncertainty and thus errors in the explanation. Indeed, we observe how non-cancerous blue areas  become smaller and turn red when inspecting the 95th percentile, whereas the cancerous areas detected remain essentially unchanged. 

Note that the presented example is intended to provide a first showcase for the potential of B-LRP in a medical use case; clearly more detailed analyses will need to follow, but they go well beyond the scope of this conceptual contribution on  XAI for Bayesian neural networks.

\subsection{Confirming the Clever Hans Effect with Bayesian LRP}

In the following experiment, we revisit the work of Lapuschkin et al. \cite{lapuschkin2016analyzing,lapuschkin2019unmasking} on \emph{clever Hans}. A \emph{clever Hans} strategy denotes a problematic solution strategy that  provides the right answer for the wrong reason: the classic example being the one of the horse Hans, which was able to correctly provide answers to simple computation questions while actually not doing math but rather reading its master\footnote{
\url{https://en.wikipedia.org/wiki/Clever_Hans}}. A modern machine-learning example is an artifact or a watermark in the data that happens to be present in one class, i.e., there is a random artifactual correlation that the model systematically and erroneously harvests \cite{lapuschkin2016analyzing,lapuschkin2019unmasking}. 
If a given artifact indeed is of high relevance for the classifier to decide for a specific class, then the artifact should be visible in a low-percentile explanation. 
We conduct a similar experiment training a VGG16 DNN on the Pascal VOC 2007 challenge dataset experiment as in \cite{lapuschkin2016analyzing,lapuschkin2019unmasking}. For the explanation of the image given in Figure \ref{fig:expCleverHans} (top right) with respect to the class \textit{horse}, we indeed observe that the watermark in the bottom left corner of the image occurs with high relevance in the 5th percentile explanation.
Hence, B-LRP enables us to confirm the clever Hans effect in our network, in the sense that the classifier draws information from artifacts that exist in the training (and also in validation and test) data set.

\section{Concluding discussion}
We proposed a novel framework for explaining and interpreting the decision making process of a Bayesian neural network. Our presented method hands two major novelties to the field of XAI: (1) it can---for the first time---explain Bayesian neural networks and (2) we can produce certainty percentiles of explanations, which allows to gauge the amount of certainty required in a given application. 
Namely, our model, B-LRP, allows to  not only  inspect the most relevant pixels for a decision, but also their (un)certainties -- a formidable starting point for obtaining novel insight into the behaviour of learning models. 
For instance, a 5th percentile level  emphasizes only the most reliable relevant explanation pixels -- a  tool perhaps helpful for more critical application domains, such as the one briefly showcased in a digital pathology application.

The results clearly demonstrate the effectiveness of the proposed method both qualitatively (e.g., Figure~\ref{fig:expMnist}) and quantitatively (e.g., Figure~\ref{fig:expPixelFlippingMNIST}).
Our quantitative evaluation on MNIST showed that B-LRP using the 5th percentile determines, for any $x\leq12$, the $x\%$ most important pixels for explanation. Similar results were obtained by our second experiment on the Imagenet dataset. Remarkably, the high-certainty explanation (5th percentile) of an image of a castle allocates relevance only to the castle object itself, while standard LRP considers also other objects in the image. This visual evaluation is confirmed also quantitatively, where the 5th percentile explanation outperforms standard LRP in the pixel-flipping evaluation. We conducted another experiment on Imagenet, which showed similar results. 


We now briefly discuss some practicalities of B-LRP. We require a trained BNN that allows us to (approximately) sample from its posterior distribution.
Or, alternatively, we can construct a Bayesian neural network using MC dropout from a standard pre-trained network (e.g., VGG16 trained on Imagenet). This provides both the explanation distribution and the respective induced uncertainties. 
Trivially, the computational complexity of B-LRP is linear to the number of posterior samples, and 
$100$ samples
 turned out  to be sufficient 
for stably assessing the explanation uncertainty on a coarse grain
in our experiments. However, more fine-grained percentiles (e.g., 1st percentile and lower)  would require more samples. An interesting line of future work will be to study functions or combinations of the percentile-based explanation to match a desired risk profile. 
While we have demonstrated our BNN explanation framework for one particular explanation method, it can readily be used for any other explanation methods and network architectures. We leave this for future work. 





\section*{Broader Impact}
Gaining a better understanding of trained neural networks by XAI techniques is beneficial to users aiming for safe, verifiable and trustworthy machine Learning. 
Specifically, the novel possibility proposed in the ms to quantify uncertainty in explanations and to be able to set the appropriate risk level in an application will be helpful in practice. 
The proposed method does not put anyone at disadvantage. If the method would fail to deliver a good
explanation in a certain scenario, there would not be direct consequences as long as the explanation
is used for decision support rather than as an own decision entity. Because the proposed explanation
method does not train a model on its own, it also does not leverage biases in the data.

\section*{Acknowledgements}
We thank Prof. Frederick Klauschen and Miriam H\"agele for their support on the pathological experiment, and Dr. Sebastian Lapuschkin for his suggestions on XAI in general.
We specially thank Luis Augusto Weber Mercado for designing and producing the overview graphics (Figure \ref{fig:BLRP}) and Matthias Kirchler for fruitful discussions.
This work was funded 
by the German Ministry for Education and Research as BIFOLD - Berlin Institute for the Foundations of Learning and Data (ref. 01IS18025A and ref 01IS18037A), and the German Research Foundation (DFG) as Math+: Berlin Mathematics
Research Center (EXC 2046/1, project-ID: 390685689). This work was partly supported by
the Institute for Information \& Communications Technology Planning \& Evaluation (IITP)
grant funded by the Korea government (No. 2017-0-00451, No. 2017-0-01779). Marius Kloft acknowledges support by the German Research Foundation (DFG) award KL 2698/2-1 and by the Federal Ministry of Science and Education (BMBF) awards 01IS18051A, 031B0770E, and 01MK20014U.

\printbibliography
\newpage

\onecolumn
\begin{appendices}
\section*{Supplement}
\appendix

The following supplementary material contains additional detailed information about the experiments conducted in our paper. Additionally, we provide an open source PyTorch implementation of our method B-LRP
\footnote{\url{https://github.com/lapalap/B-LRP}}.

\section{Pixel-Flipping}
\label{sec:PixelFlipping}
For quantitatively evaluating our proposed method, B-LRP, we employ pixel-flipping \cite{samek2016evaluating}---a practical technique to quantify the goodness of an explanation with respect to a specific decision of a trained model.
Given relevance scores (e.g., in the form of heatmap) attributed to all pixels
by an explanation method,
pixel-flipping 
performs the following steps.  First, the pixels 
are ranked in descending order of their relevance scores, i.e., pixels with higher relevance scores are ranked first.
Then, the original pixel values are iteratively perturbed (e.g., set to zero or replaced with random values), according to the ranking. 
Namely, the pixel with $k$-th highest
relevance score is perturbed at the $k$-th iteration. 
The prediction score (output of the trained model for the perturbed input) is evaluated and recorded at each step.
Typically, we plot the resulting curve 
averaged over the test images is plotted as a function of $k$.
This allows us to compare explanation methods: the better an explanation method is, the steeper the prediction score drops,
because the most relevant pixels 
identified by a good explanation
should affect the prediction score most.

The pixel-flipping procedure is summarized in Algorithm~\ref{alg:PixelFlipping}.
\begin{algorithm}[H]
\caption{Pixel-Flipping}
\begin{algorithmic} 
\REQUIRE  ${x}$ -- input image, $R$ -- relevance, $f$ trained model
\ENSURE $scores$ -- sequence of decaying prediction scores.
\STATE $scores \leftarrow $[]
\FOR{$p$ in argsort(-$R$):}
\STATE Perturb the pixel $x_p$.
\STATE $scores$.append($f({x}, W)$).
\ENDFOR
\RETURN $scores$
\end{algorithmic}
\label{alg:PixelFlipping}
\end{algorithm}


In our experiments, we perturbed pixels by replacing it with a random variable by
$$x_p \leftarrow \frac{U - \mu}{\sigma},$$  where $x_p$ denotes the $p$-th pixel of the image $x$, $U \sim \textrm{Unif}[0,1]$ is a random variable following the uniform distribution, and $\mu$ and $\sigma$ are the mean and the standard deviation, respectively, of the corresponding pixel values over the training set. 


\section{Visualization Details of LRP and B-LRP Explanations}

For visualization, 
each Standard LRP and B-LRP explanation is normalised using the MinMax transformation \cite{bach2015pixel},
which maps positive relevances to $[0,1]$ and negative relevances to $[-1,0]$. Namely, the positive relevances are divided by the maximal positive relevance over the pixels, and the negative relevances are divided by the absolute value of the minimal negative relevance over the pixels. 
We use the \textit{'seismic'} colormap\footnote{\url{https://matplotlib.org/3.1.0/tutorials/colors/colormaps.html}}, which attributes red tones to pixels with positive relevances and blue tones to the pixels with negative relevances.
\begin{figure}[h]
\centering
 \includegraphics[width=0.3
 \textwidth]{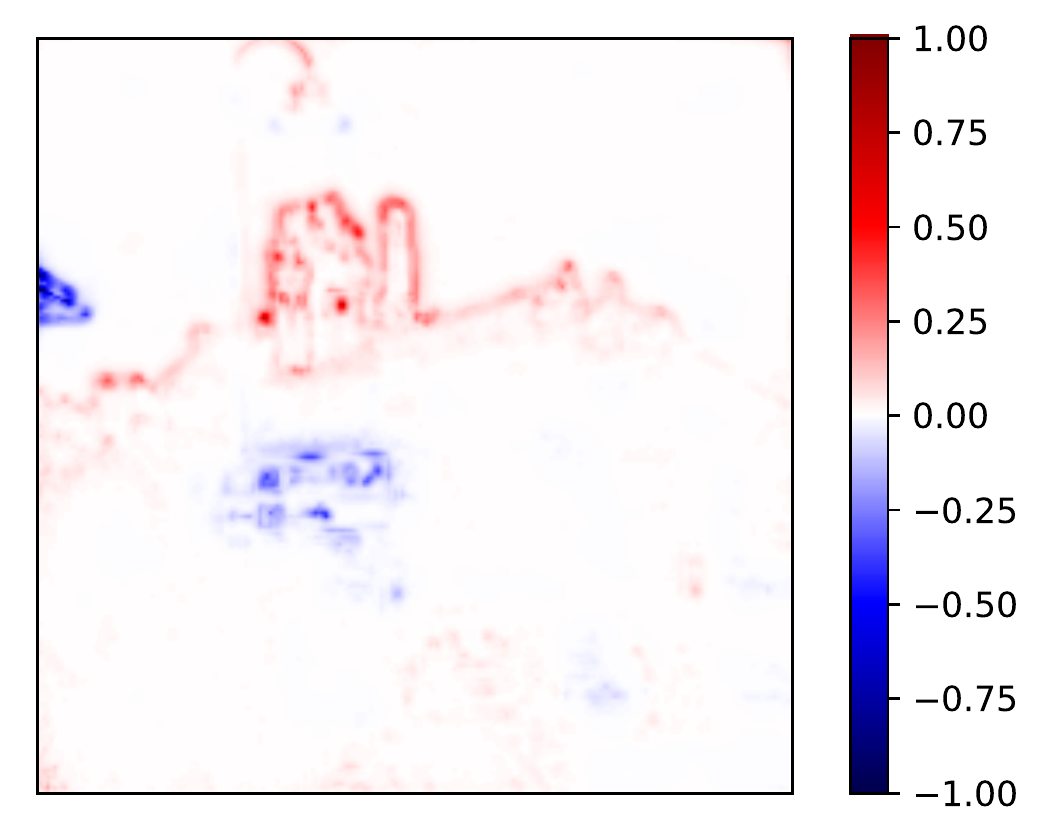}
\caption{Visualization of the relevances for the predicted class \emph{castle} found by the LRP-CMP rule. The colorbar on the right depicts the color range between [-1,1].
\label{fig:appendix_castle_colorbar}}
\end{figure}
\section{Experiment on MNIST Data}
\label{sec:MNISTDetails}
\begin{figure}[h]
\centering
 \includegraphics[width=
  \textwidth]{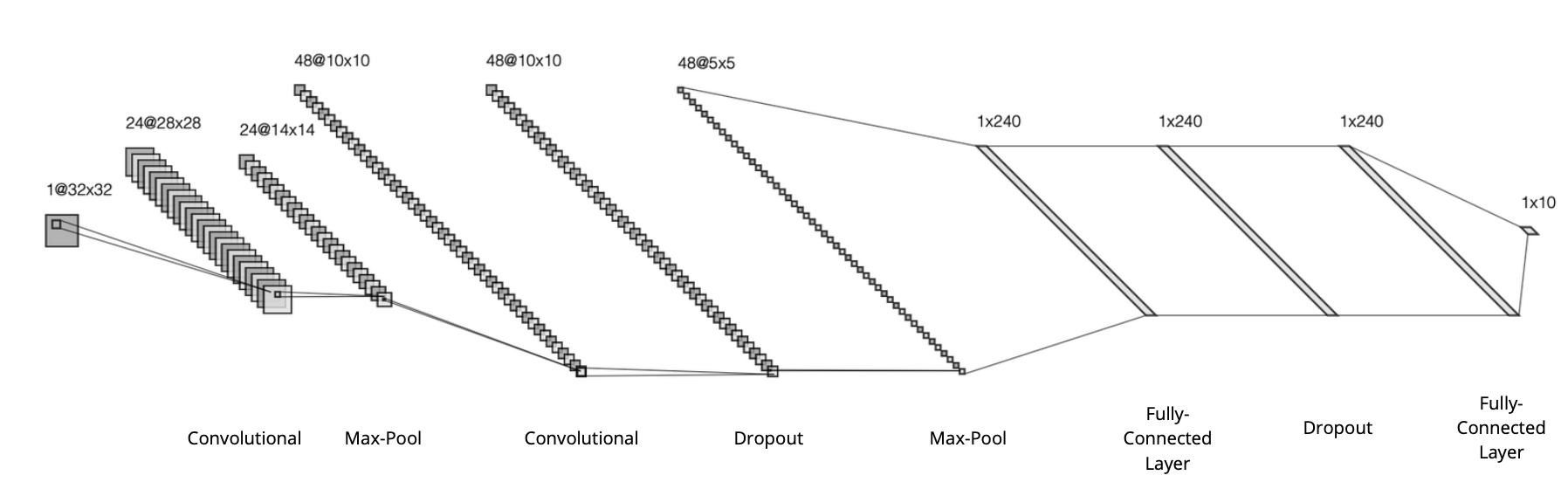}
\caption{Network architecture used for the MNIST experiment. 
\label{fig:appendix_lenet_mnist}}
\end{figure}

%
For the experiment on the MNIST data set, we employed a feed-forward convolutional neural network, similar to LeNet, with additional dropout layers. The architecture is as follows: Convolution2d (24 neurons, kernel size = 5), ReLU, MaxPool (kernel size = 2),  Convolution2d  (48 neurons, kernel size = 5), ReLU, Dropout (p = 0.5),  MaxPool (kernel size = 2), 
Fully-Connected (240 neurons), ReLU, Dropout (p = 0.5), Fully-Connected (10 neurons).
Figure~\ref{fig:appendix_lenet_mnist} visualizes the architecture.

We trained the network on 
the 50,000 images in the MNIST training set.
The images were pre-proceessed (rescaled) so that the mean and the standard deviation of each pixel over all training images are 0 and 1, respectively.
We minimized the standard cross-entropy loss by Stochastic Gradient Descent (SGD)
with batch size $32$, where the learning rate and the momentum were set to $0.001$ and $0.9$ respectively. The network was trained for 50 epochs, reaching 99.1\% accuracy on the 10,000 images in the MNIST test set.
%
\section{Experiment on Histopathological Data}
For the experiment on the Haematoxylin-eosin-stained Lung adeno carcinoma (LUAD) (and non cancer slices) data, we used a standard VGG19 model, where the number of output neurons in the output layer was adjusted to two corresponding to `cancer' and `non-cancer'. All layers, except for the last one were initialised by a weights from a pretrained VGG19 on Imagenet. For training procedure and parameter setting, we followed
\cite{hagele2020resolving}. 
The trained network achieves 
the weighted $F_1$-score
$0.9047$ on the test set.

In medical applications, it is common to apply a specific transformation to the relevance scores for visualization,
in order to make it visually more accessible to experts.
Specifically, 
we applied the following monotonic transformation to the relevance score: 


$$ T(R) = \sqrt{|R|} \cdot \sign \left (R\right),$$
where each operator applies pixel-wise.
This transformation makes weak relevances more visible,
while keeping the ranking of relevance scores.

\section{Experiment on Clever Hans Effects}

For the Pascal VOC 2007 multi-label classification experiment, we employed a standard VGG16 network \cite{simonyan2014very},
and adjusted the number of output neurons from 1000 to 20, which is the number of different classes in the Pascal VOC dataset.

We resized 
each training image
so that the shorter axis has 224 pixels, keeping the aspect ratio  unchanged.  Then, we randomly cropped the longer axis, and obtained square images with the size 224$\times$224. 
We trained the network for 60 epochs, by minimizing the Binary Cross Entropy loss preceded with a Sigmoid layer\footnote{\url{https://pytorch.org/docs/master/generated/torch.nn.BCEWithLogitsLoss.html}}.
We used the Adam optimiser with its parameters set to $\alpha = 0.0001, \beta_1 = 0.9, \beta_2 = 0.999$. Our trained VGG16 network achieves 91.6\%\footnote{\url{https://scikit-learn.org/stable/modules/generated/sklearn.metrics.accuracy_score.html}} in the multi-label classification on the test set,
for which center cropping with square size of 224$\times$224 was applied, instead of random cropping.

\begin{figure}[H]

\includegraphics[width= \textwidth]{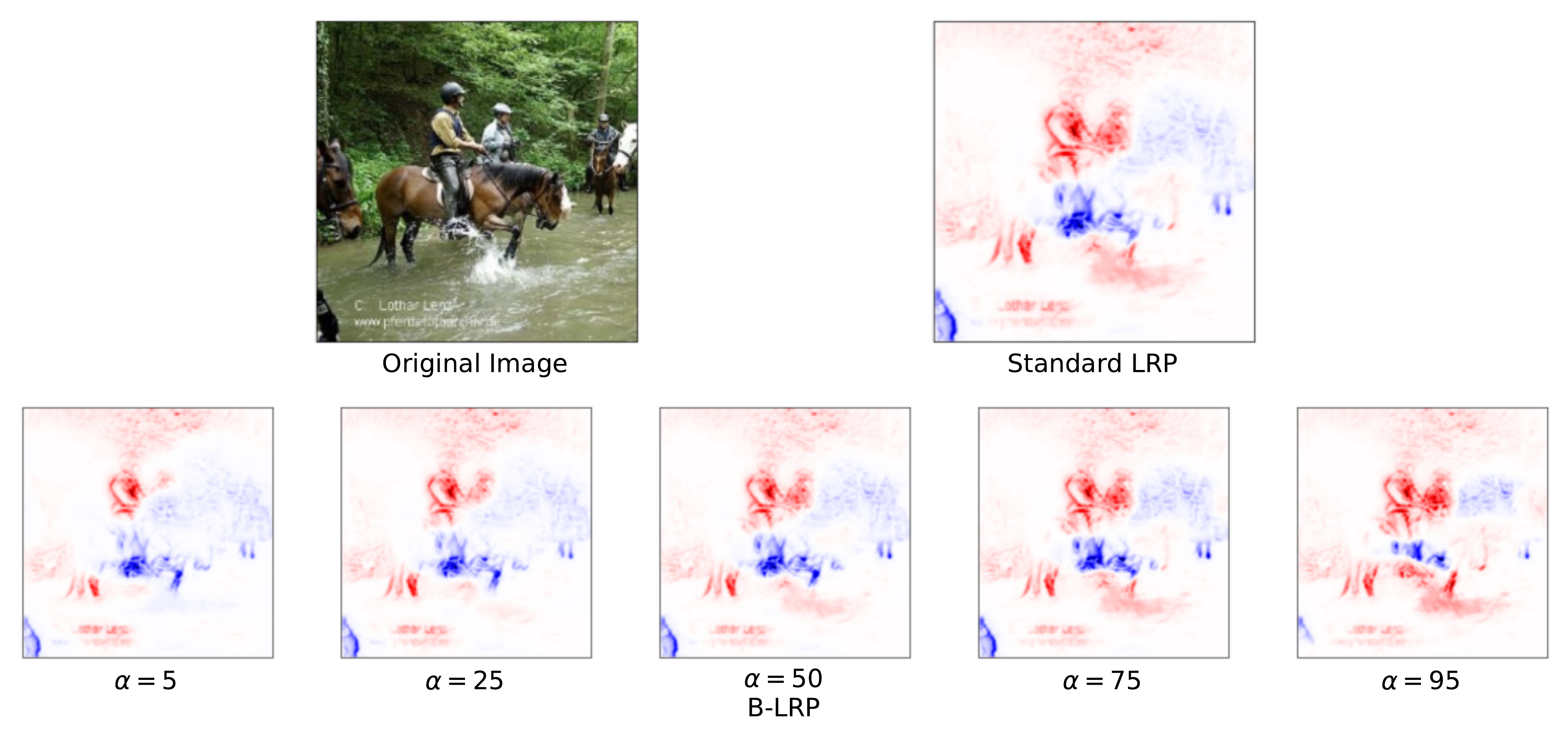}
\vspace{-0.5cm}
\caption{Another exemplary visualization of the \emph{clever hans} effect for an image depicting a horse.}
\end{figure}
\end{appendices}

\newpage

\end{document}